\documentclass[letterpaper, 10 pt, conference]{ieeeconf}  

\IEEEoverridecommandlockouts                              

\usepackage{graphics} 
\usepackage{epsfig} 
\usepackage{mathptmx} 
\usepackage{times} 
\usepackage{amsmath} 
\usepackage{amssymb}  
\usepackage{graphicx}
\usepackage{subfigure}
\usepackage{multirow}
\usepackage{hyperref}
\usepackage{xcolor}

\title{\LARGE \bf
A Method to Generate High Precision Mesh Model and RGB-D Dataset for 6D Pose Estimation Task
}

\author{Minglei Lu$^{1}$, Yu Guo$^{2}$, Fei Wang$^{1}$, Zheng Dang$^{1}$ 
\thanks{$^{1}$Institute of Artificial Intelligence and Robotics, Xi’an Jiaotong     University, China
{\tt\small am12345@stu.xjtu.edu.cn, wfx@mail.xjtu.edu.cn, dangzheng713@stu.xjtu.edu.cn}}
\thanks{$^{2}$School of Software Engineering, Xi'an Jiaotong University, China {\tt\small yu.guo@xjtu.edu.cn}}%
}
\begin{document}
\maketitle
\thispagestyle{empty}
\pagestyle{empty}
\newcommand{\argmin}{\mathop{\mathrm{argmin}}}

\newcommand{\GY}[1]{{\color{red}{\bf GY: #1}}}
\newcommand{\gy}[1]{{\color{red} #1}}
\newcommand{\ZD}[1]{{\color{blue}{\bf ZD: #1}}}
\newcommand{\zd}[1]{{\color{blue}{#1}}}
\newcommand{\LML}[1]{{\color{green}{\bf LML: #1}}}
\newcommand{\lml}[1]{{\color{green} #1}}

\begin{abstract}

Recently, 3D version  has been improved greatly due to the development of deep neural networks. A high quality dataset is important to the deep learning method. Existing datasets for 3D vision has been constructed, such as Bigbird and YCB. However, the depth sensors used to make these datasets are out of date, which made the resolution and accuracy of the datasets cannot full fill the higher standards of demand. Although the equipment and technology got better, but no one was trying to collect new and better dataset. Here we are trying to fill that gap. To this end, we propose a new method for object reconstruction, which takes into account the speed, accuracy and robustness. Our method could be used to produce large dataset with better and more accurate annotation. More importantly, our data is more close to the rendering data, which shrinking the gap between the real data and synthetic data further.
\end{abstract}
\section{INTRODUCTION}
The quality of the dataset usually determines the upper limit of the accuracy that the algorithm can achieve. Using more accurate annotated dataset, the accuracy could be further improved has been validated in the paper~\cite{Grenzdorffer20}.  Thus making a high precision dataset is not a trivial task. Reconstruction is traditional topic in computer vision, which has been widely used in many applications, such as robot manipulation~\cite{Hashimoto11}, pose estimation~\cite{Hodan20} and augmented reality~\cite{Carmigniani11}.
With the development of deep learning method and pose estimation, the need for highly accurate dataset is becoming more and more urgent.

At the same time, depth sensor technology has also been developed.
Primesense-based RGB-D cameras like the Microsoft Kinect~\cite{Zhang12} and Intel RealSense~\cite{Draelos15} is the first generation of depth sensors, and have been gradually replaced by higher precision and resolution cameras such as Orbbec
Astra, Basler ToF, TENGJU.


There are several paper~\cite{Singh14,Calli15} which also give the way to reconstruct objects' mesh and genereate RGB-D data. But these paper have been written in 4 years ago, and the devices and methods used could not be considered as the state of the art method today. To this end, we redesign the pipeline, adjust the equipment setting and give a new dataset capturing solution using the commercial equipment available today.

\section{RELATED WORK}

\begin{figure*}[ht]
    \centering
    \subfigure{\includegraphics[width=0.21\linewidth]{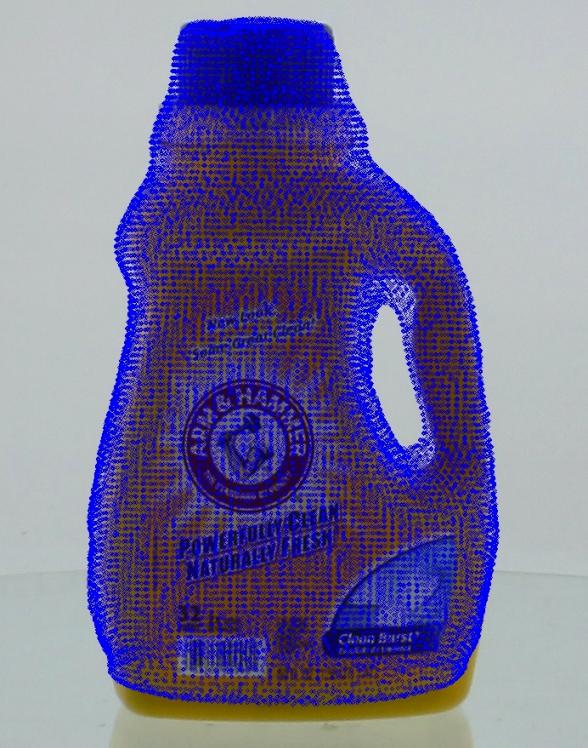}
    \includegraphics[width=0.21\linewidth]{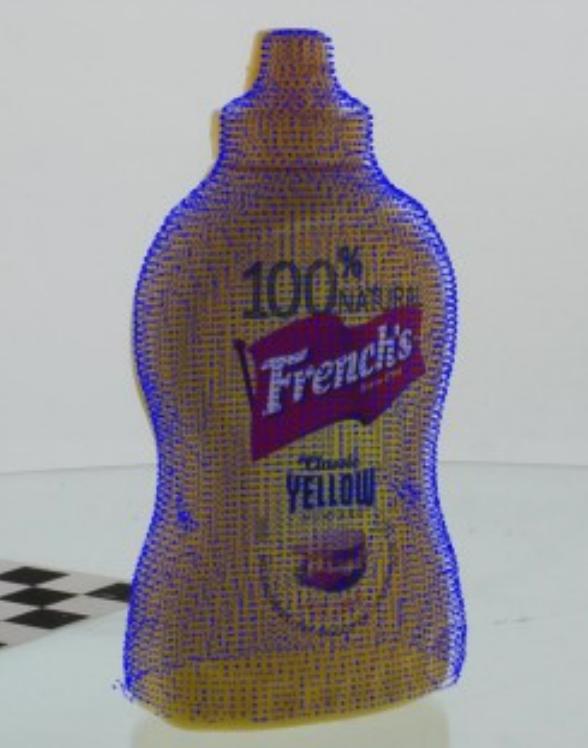}
    \includegraphics[width=0.21\linewidth]{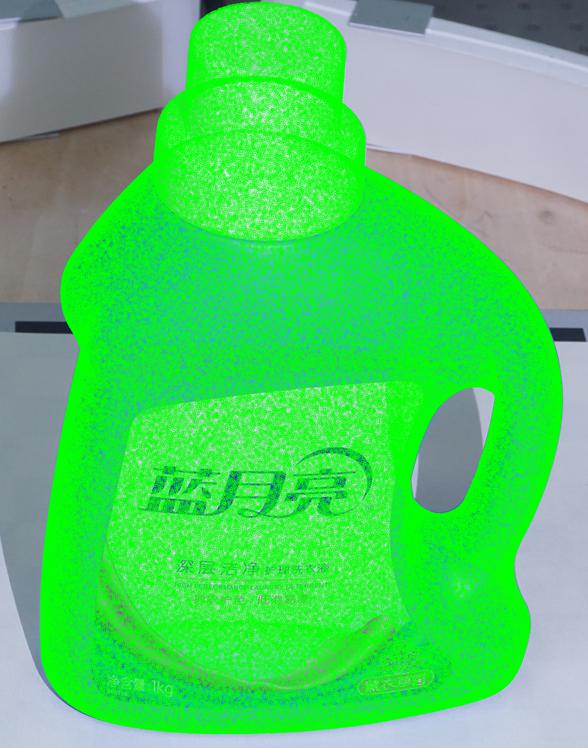}}
    \caption{The left image shows Bigbird's outcome, the middle image shows YCB's outcome and the right image shows ours(in order to have a better view, we only sample one tenth of the points for projection). Our method achieves pixel level accuracy, and the details of the object's edge is better.}
    \label{projection}
\end{figure*}
\begin{figure*}[ht]
    \centering
    \subfigure[Xtion]{\includegraphics[width=0.18\hsize]{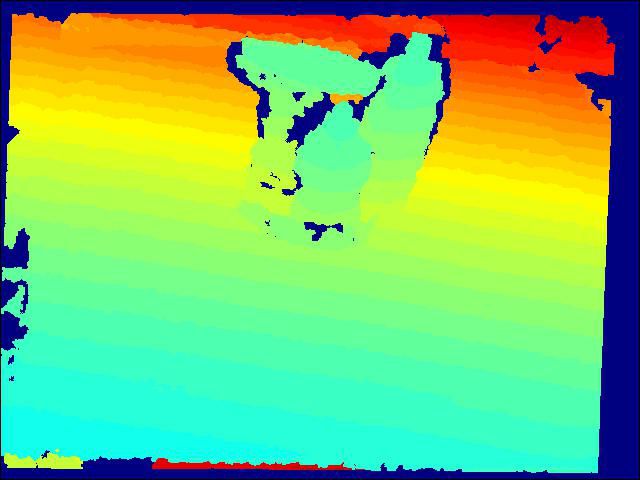}}  
    \subfigure[Astra]{\includegraphics[width=0.18\hsize]{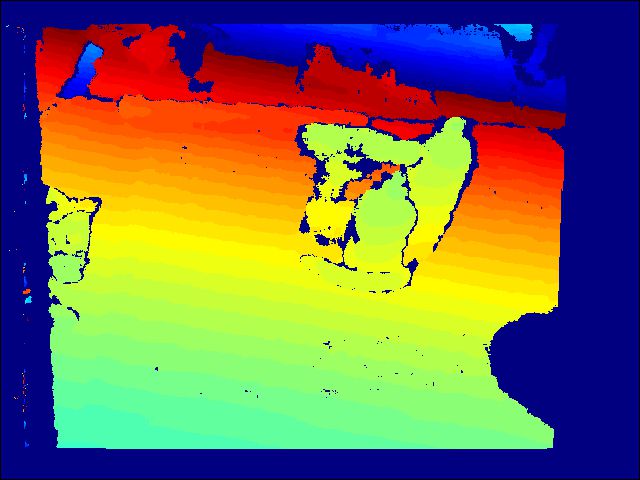}} 
    \subfigure[Kinect2]{\includegraphics[width=0.18\hsize]{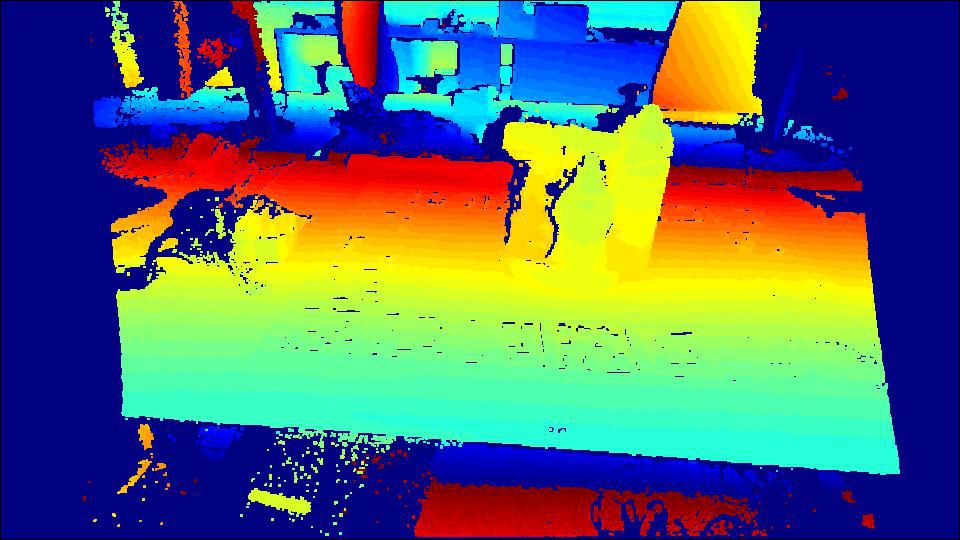}}
    \subfigure[Realsense R200]{\includegraphics[width=0.18\hsize]{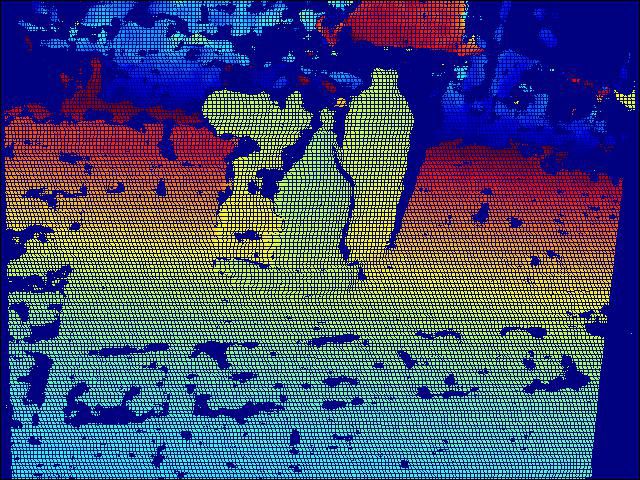}}
    \subfigure[Basler ToF]{\includegraphics[width=0.18\hsize]{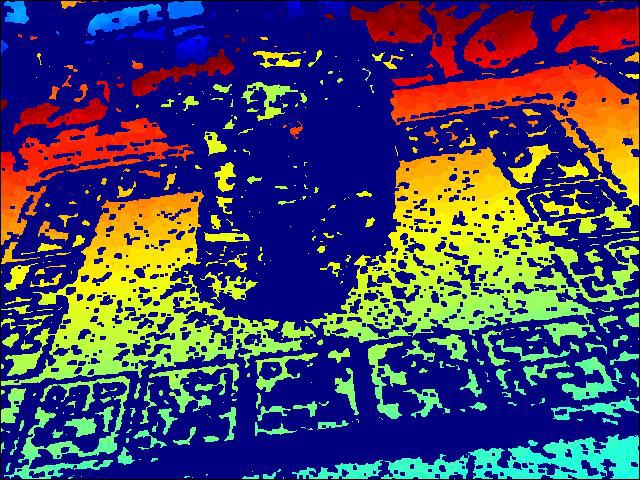}}
    
    \subfigure[pico flexx]{\includegraphics[width=0.18\hsize]{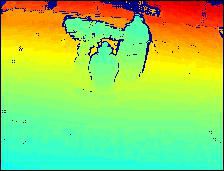}}
    \subfigure[Ensenso N35]{\includegraphics[width=0.18\hsize]{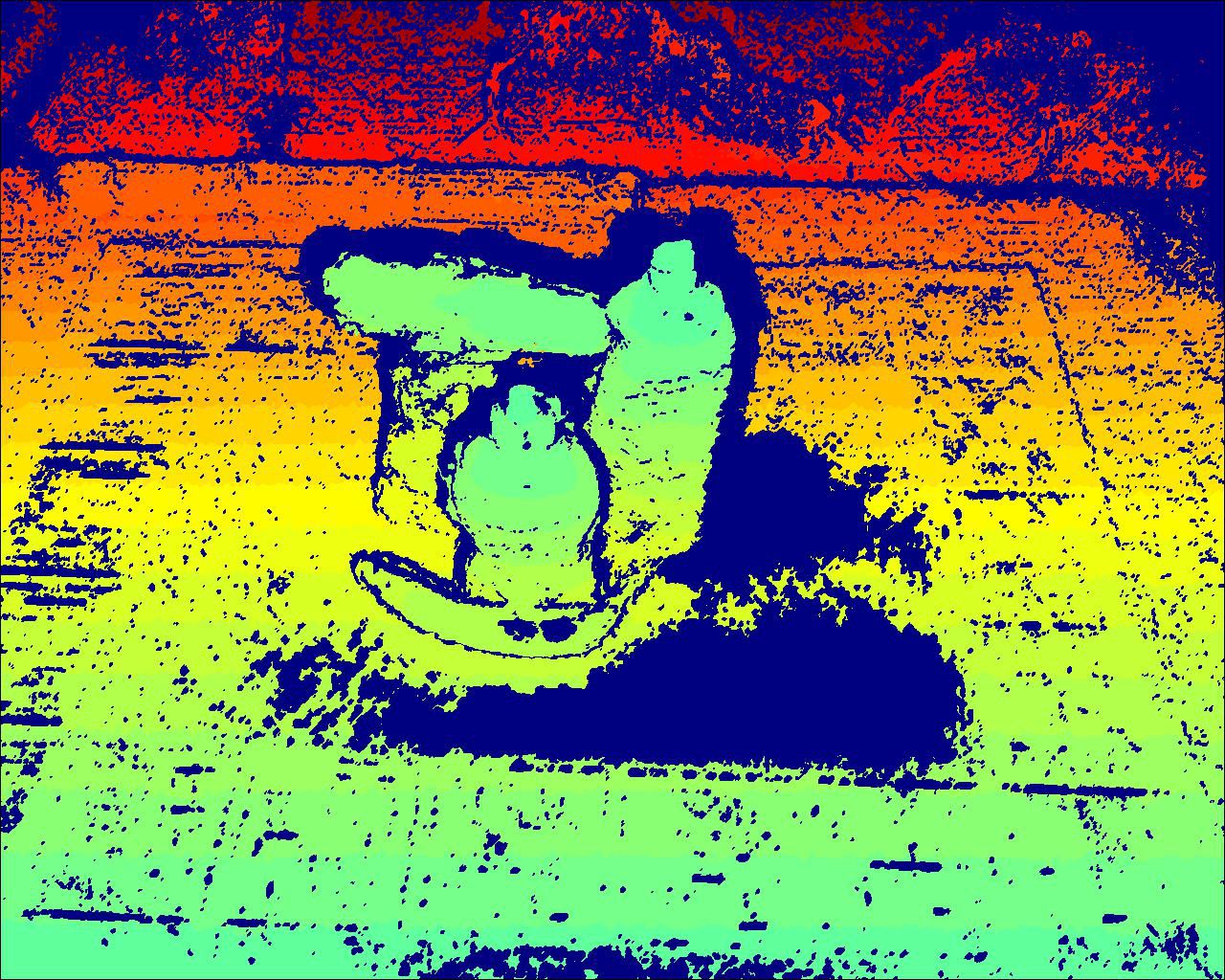}}
    \subfigure[Ours]{\includegraphics[width=0.18\hsize]{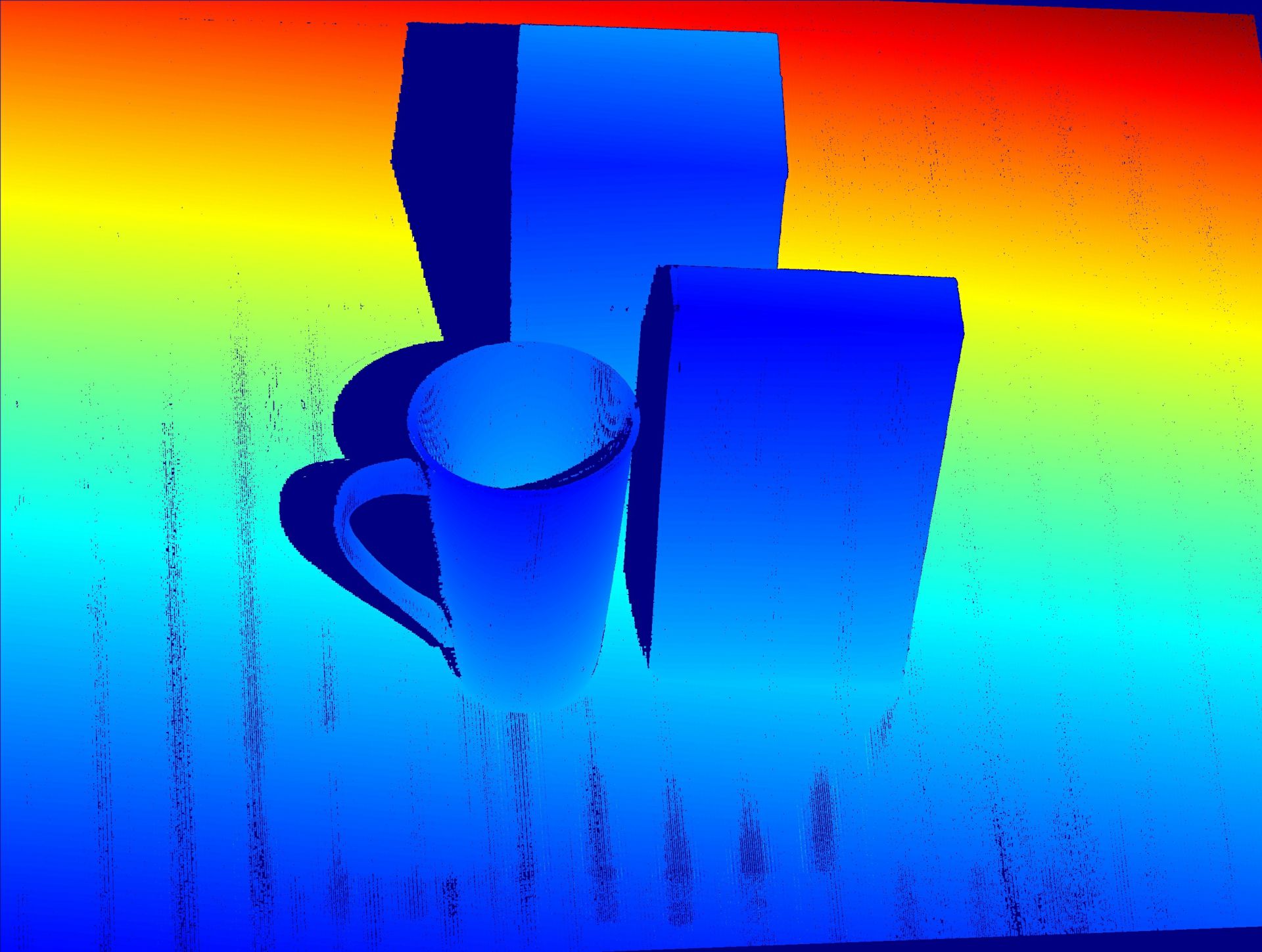}} 
    \subfigure[Render]{\includegraphics[width=0.18\hsize]{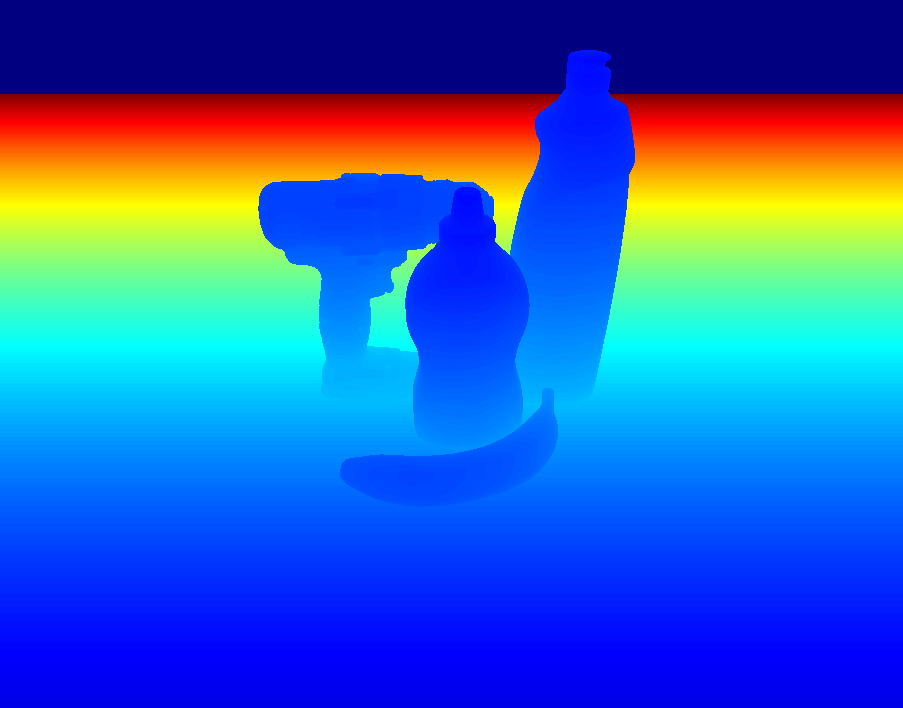}}
        
    \caption{Depth maps generated by several depth sensors(mentioned in YCB-M and ours) and by render method. Compared with depth map resolution in YCB (up to 1.3MP), ours has higher resolution(6MP) and the edge of object is sharper.}
    \label{depth}
\end{figure*}

In the recent years, many mesh datasets of objects has been provided \cite{Kasper12, Hinterstoisser12, Calli15} of objects. They are widely used in various fields, such as 6D pose estimation \cite{Xiang18, Hu19, Wang19f} and point cloud registration \cite{Gojcic20, Choy2020, Yew20}. 

\subsection{Data Collection}
To create a dataset, a method's speed, accuracy, robustness and pipeline automation should all be considered. These affects are highly influenced by the performance of system and devices.
Kinect is the first generation of the commercial depth sensor, the resolution of its depth map is $640 \times 480$, the error is more than 1mm~\cite{Zennaro15}.
By using multiple sensors and calibrating between multiple color and depth sensors \cite{Herrera11, Geiger12}, Bigbird~\cite{Singh14} can construct a 3D reconstruction system and \cite{Calli15} followed this method. The reprojection results by using this method shows than this method can successfully used for 3D reconstruction. But The details of the object are not well handled, and they did not provide a effective method to reconstruct the bottom of objects~\cite{Calli15}. 

YCB-M~\cite{Grenzdorffer20} compared several depth sensors spanning three depth sensing technology (active stereo, time-of-flight and structured light). They only use these sensors in collecting data in new scenes and marking ground truth, but the 3D object mesh models are still the old one. 
The quality of 3D model will also affect the performance of the algorithm, especially for the deep learning methods using synthetic data for training.

Recently there is a tendency people trying to use the synthetic for training, but evaluated on the real scene dataset~\cite{Park19a, Wada20, Wang19e}. Due to the huge gap of between the real scene and the synthetic data, these algorithm usually doesn't work well on the real scene data~\cite{Wang19e}.

Structured light sensing technology has been developed \cite{Scharstein03} \cite{Wang12} in these years, the accuracy and resolution of structured light camera have been greatly improved. Structured light camera can provide high quality point clouds and depth maps. 

\subsection{Contributions}
Take advantage of the advances made by various devices and algorithms, we propose a fast high precision 3D reconstruction method. Our contributions can be summarized as follows:
\begin{itemize}
   \item [1)] We present a robust, high precision and fast 3D reconstruction method, and we can reconstruct the bottom scene of objects which is not mentioned in other methods of making data sets. 
\end{itemize}
\begin{itemize}
    \item [2)] Our method can give more accurate annotation data, can be used to produce datasets with more accurate ground truth.
\end{itemize}
\begin{itemize}
    \item [3)] Our data is closer to the datasets generated by render, reducing the gap between real data and synthetic data.
\end{itemize}
\begin{itemize}
    \item [4)] 
    We've given a couple of methods, and they all improve the results in our experiment.
\end{itemize}

\section{SYSTEM OVERVIEW}
\begin{figure}[htbp]
    \centering
    \includegraphics[scale=0.18]{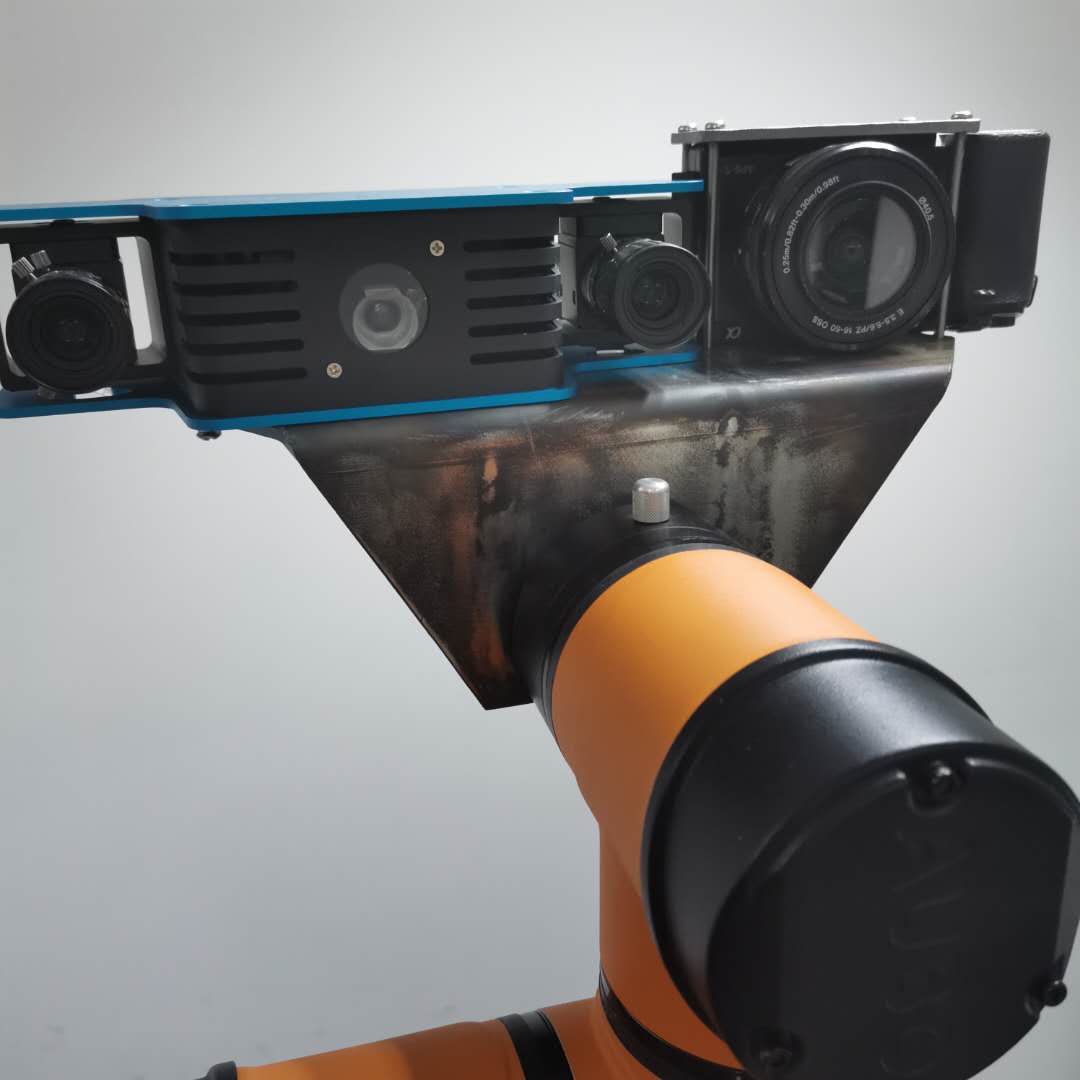}
    \caption{Two cameras connected together using mount}
    \label{camera}
\end{figure}
\begin{figure}[htbp]
    \centering
    \includegraphics[scale=0.2]{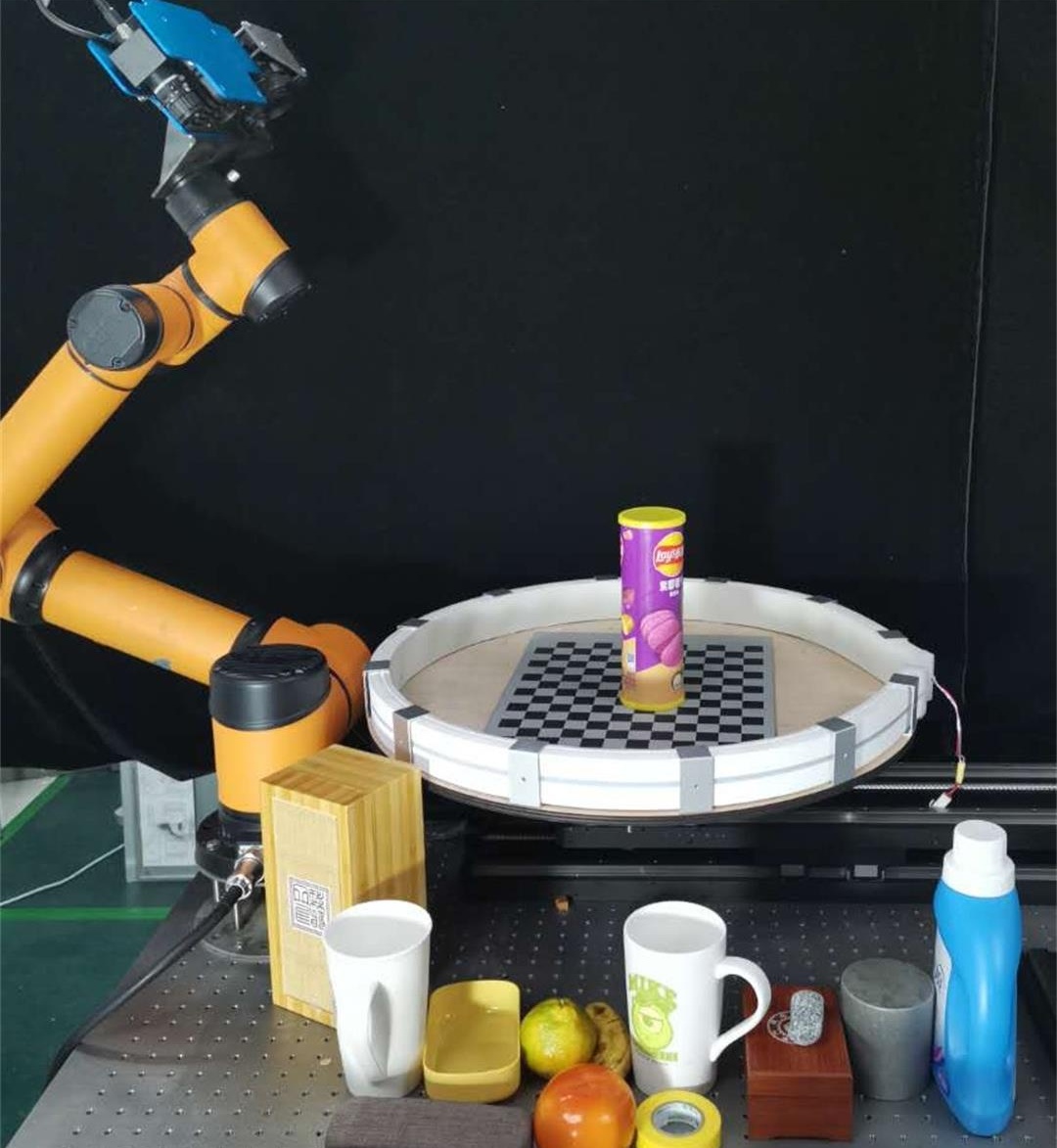}
    \caption{Side view of our system.We put the object on the center of chessboard and run the program to achieve automatic acquisition.}
    \label{system}
\end{figure}
The sensors in our system comprise of one high resolution SONY ILCE-6000 camera($6000\times4000$ resolution) and one TENGJU DH060 structured light camera($3072\times2048$ resolution). These two cameras are mounted to the end effector of a AUBO robot arm, as shown in Fig.~\ref{camera}. The object would be placed on a high-precision 7SC404-Z turntable. The turntable carries a tray with a diameter of 600mm. Around the tray, a LED light band is attached to the edge of it. In order to obtain the calibration data and accurately estimate the pose, a chessboard is placed at the cameras' center of view.

\begin{figure}[htb]
    \centering
    \subfigure{\includegraphics[width=0.45\linewidth]{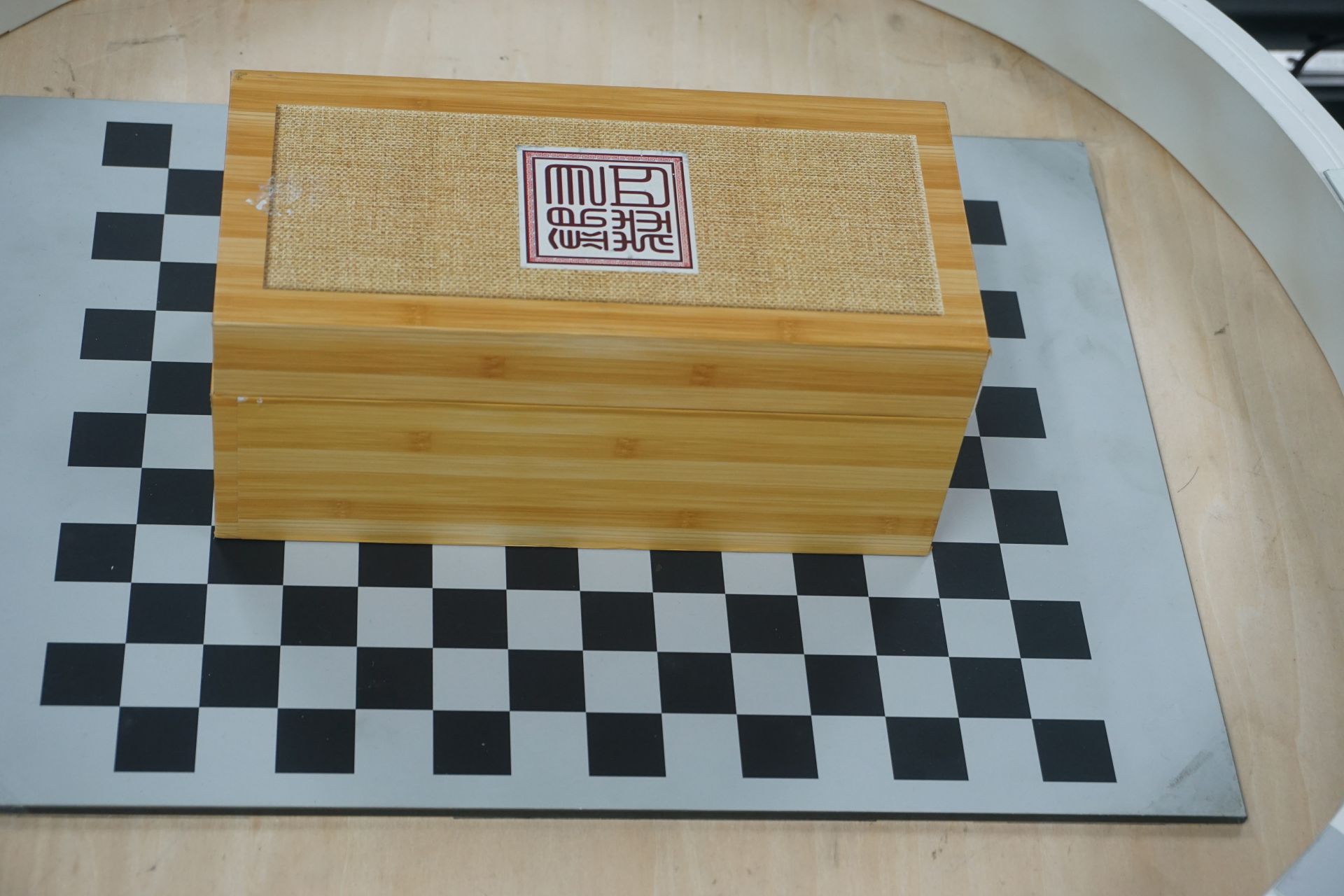} \includegraphics[width=0.45\linewidth]{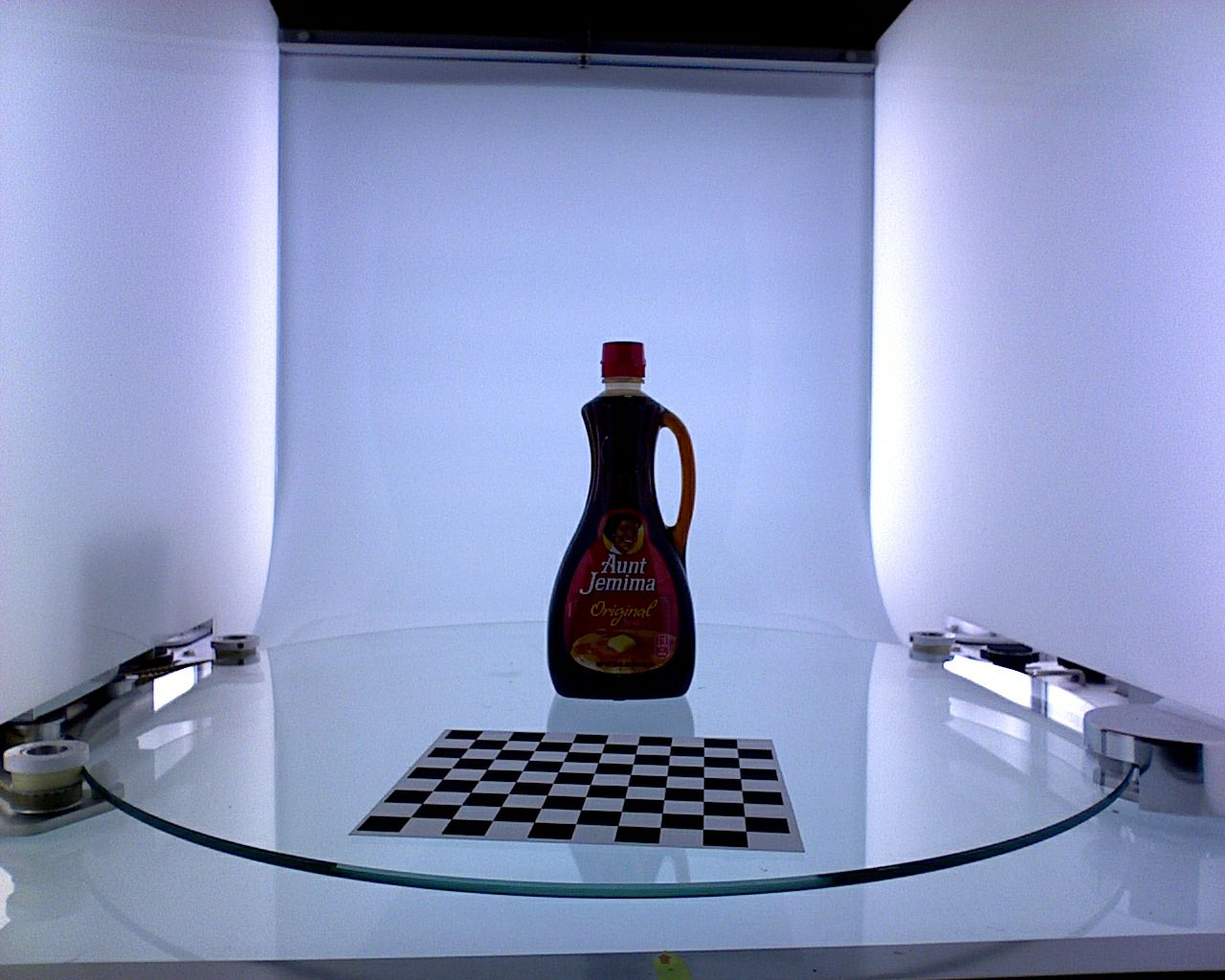}}
    \caption{Comparison of our picture and Bigbird picture.The left is ours and we can see the object takes much more pixels than Bigbird's picture in right, this makes object clearer in RGB image and gets more dense point cloud in one scene.}
    \label{picture compare}
 \end{figure}

For the whole reconstruction process, we only calibrate the pose between cameras and the turntable one and only one time at the beginning. No matter capturing how many objects we keep using the same pose for reconstruction. To be specific, we rotate the turntable in increments of 22.5 degrees and place the robot arm at 2 positions to get the images of the chessboard. Then these images would be used to calculate the pose of the cameras. These calibrated camera poses would be used for the next process. After the calibration process, we place the object on the turntable and rerun the process to get the images of the object. From our experience, if the chessboard been occluded, normally leading to inferior at the reconstruction accuracy. Thanks to the high repositioning accuracy of the turntable and the robot arm, we could reuse the calibrated camera pose for the reconstruction process. This split capturing method gives us an advantage that we don't have to worry about the chessboard been occluded by the object in the reconstruction process, which could be happened in the paper Bigbird~\cite{Singh14} since they capture the object and chessboard at the same time~\ref{picture compare}. 
We found two observation positions of the camera is enough to capture all the structure information of the object. The first position is placing the camera above the object and looking at the object. The second position is placing the camera at a level with the center of the object and looking at the center of the object.After the image capturing process, yielding 32 point clouds from the structured light camera and 32 high-resolution RGB images from the Sony camera which would be used for the following process.


For a single object, the process of data collection is as follows:
\begin{itemize}
    \item [1)]
    Start all cameras and robot arm and put one object at the center of the turntable.
\end{itemize}
\begin{itemize}
    \item [2)]
    Capture object images at two positions of the manipulator. For each location of manipulator, The turntable rotates to 16 angles. For each manipulator position, the process is as follows:
    \begin{itemize}
        \item [a)]
        Sony camera capture an image in 1/20s.
    \end{itemize}
    \begin{itemize}
        \item [b)]
        The structured light camera turns on the structured light and captures 12 images for each eye at a speed of 10FPS, then the structured light camera turns off structured light and captures one RGB image. This step costs 1.9s.
    \end{itemize}
    \begin{itemize}
        \item [c)]
        The structured light camera generates a raw point cloud from 24 images with structured light and the turntable rotates $22.5^{\circ}$ at the same time. 
        Rotating turntable and generating a raw point cloud will operate in parallel which will takes 5.6s. When turntable rotated $360^{\circ}$, the robot arm moves to the next location. This process also operate simultaneously with previous two steps.
    \end{itemize}
\end{itemize}

In order to capture the bottom structure and texture of the object, we reverse the object on the table and take a second shot of the object. The whole process costs about 7min.
\section{CALIBRATION}
As the camera lenses of the structured light camera are industrial lens with low resolution and obvious color distortion, we use Sony camera to redraw color on the raw point cloud. 
The raw point cloud is defined under the camera's left eye coordinate system. In order to align the raw point cloud and RGB image, we need to get the relative transformation between the structured light camera and RGB camera. We made a iron joint to fixed these two cameras, to make sure these two camera's relative position won't change during the data collection procedure. For this reason, the extrinsic matrix between these two cameras only need to be calibrate once.

\subsection{Calibration between RGB camera and Depth camera}


In the calibration process, we capture n (n=75 in practice) chessboard images and use OpenCV's camera calibration package~\cite{Bradski08} which based on the method~\cite{Zhang99} to calibrate both intrinsic and extrinsic parameters. 
we define $\Gamma_{rgb}$ as the set of RGB camera's extrinsic matrix and $\Gamma_{depth}$ as the set of depth camera's extrinsic matrix. $\Gamma_{rgb}=\{T_i\in R^{4 \times 4}:T_1, \cdots, T_n\}$ and $\Gamma_{depth}=\{T_i'\in R^{4 \times 4}:T_1', \cdots, T_n'\}$, we use these two notions in the next step.

From each scene, we can calculate the transformation between two cameras using the extrinsic matrix of two cameras. Due to the fact that these two cameras have been fixed to the end effector of the robot arm, there will not have any relative movement during the data acquisition process. Theoretically, no matter we use which pair of the extrisic matrices to calculate the $t_{trans}$ should have the same result. However, due to the noise, the calculated value in different scenarios will have a small residual, and the value is normally distributed around the ground truth. Thus we construct an optimization problem to refine the transformation matrix $T_{relative}$ between RGB camera and depth camera.

To this end, for scene $i$, the transformation between two cameras is $T_i''=T_i'T_i^{-1}$, $T_i \in \Gamma_{rgb}$ and $T_i' \in \Gamma_{depth}$. We define $\Gamma_{relative}=\{T_i''\in R^{4 \times 4}:T_1'', T_2'', \cdots,T_n''\}$ as the set of transformations between two cameras. The loss function is given by:


\begin{equation}
    T_{relative} = \argmin_{t}\sum_{i=1}^n\left\|T_i''-t\right\|_1
\end{equation}

where $T_{relative}$ denotes the final relative transformation matrix between RGB camera and depth camera.

\subsection{Calibration for each scene}
\label{bound}

\begin{figure}[htbp]
    \centering
    \includegraphics[scale=0.09]{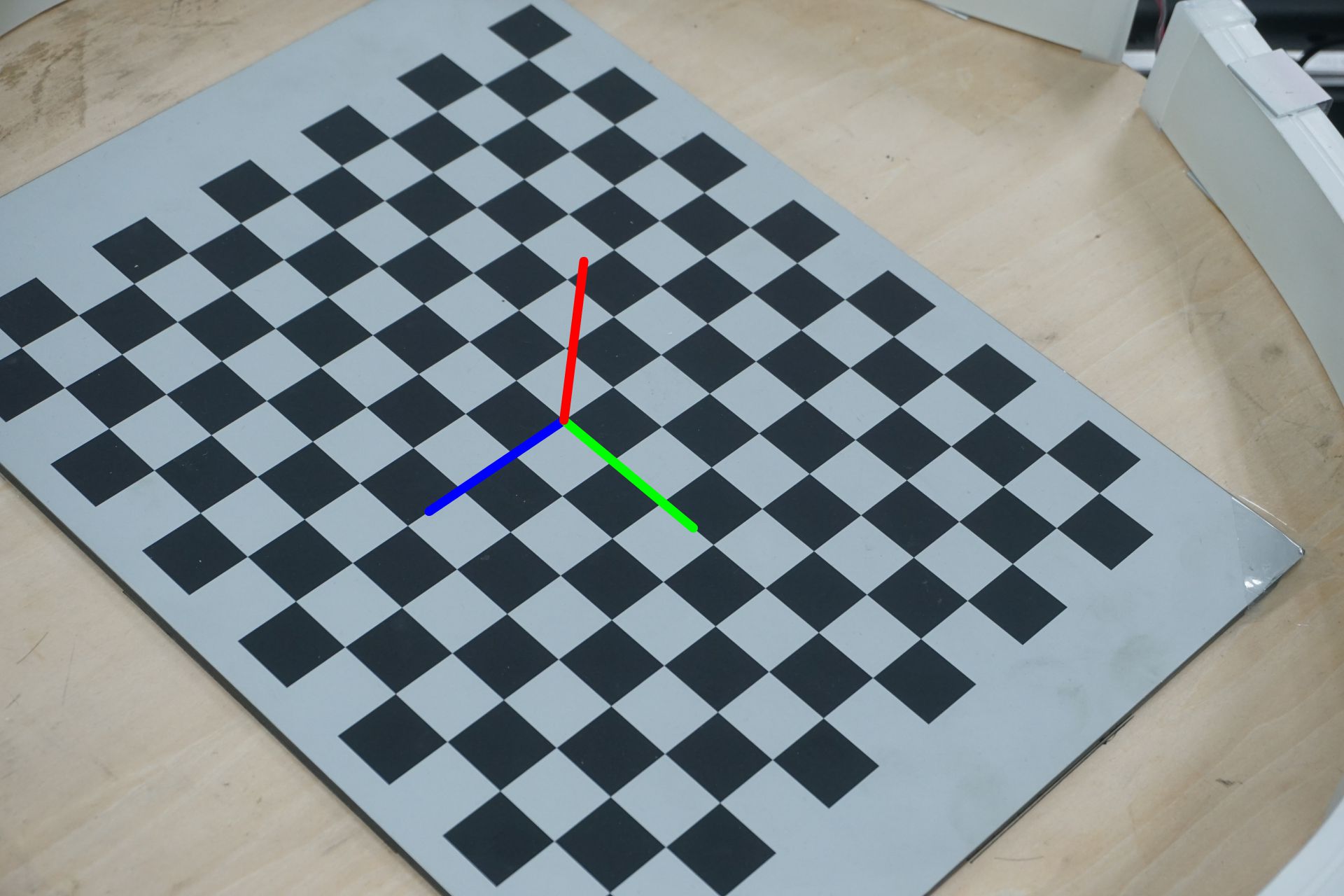}
    \caption{This picture shows the reference coordinate, the origin is the center of chessboard corners, the The blue, green and red axis represent X , Y and Z axis respectively.}
    \label{chessboard}
\end{figure}

\begin{figure}[htb]
    \centering
    \includegraphics[scale=0.09]{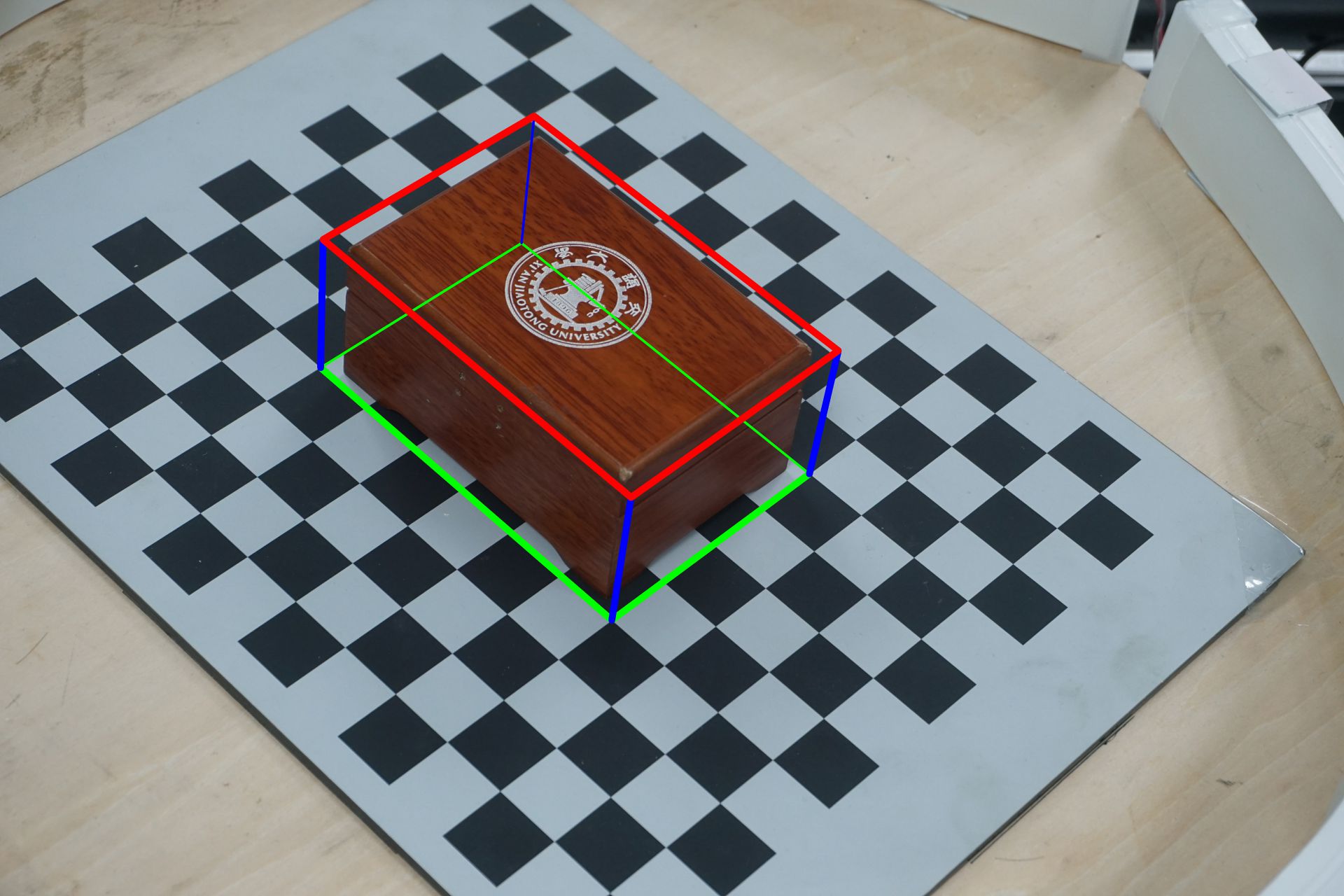}
    \caption{We can give each object a rough bounding box for segmentation.}
    \label{boundingbox}
\end{figure}

For each turntable angle and manipulator position, we can calibrate the transformation matrix for this scene. 
The transformation matrix represents the relative transformation between the depth camera coordinate and the reference coordinate~\ref{chessboard}, which could be calculate by using checkerboard and solvePnP~\cite{Penate2013}.
By transforming point clouds from camera coordinates to the reference coordinate, we realize point cloud fusion under multi-view. We can define a rough bounding box for each object. We put all objects in the center of the chessboard. We can estimate the length and width of the bounding box by using corners of the chessboard, and give a rough estimation of the height with a ruler. The bounding box is used to segment objects from point clouds(because of the high quality of point clouds generated by the structured light camera, we do not need to be very strict in bounding box's segmentation), as shown in Fig.~\ref{boundingbox}.
\section{Mesh Generation}

\begin{figure}[htbp]
    \centering
    \subfigure{\includegraphics[width=0.45\linewidth]{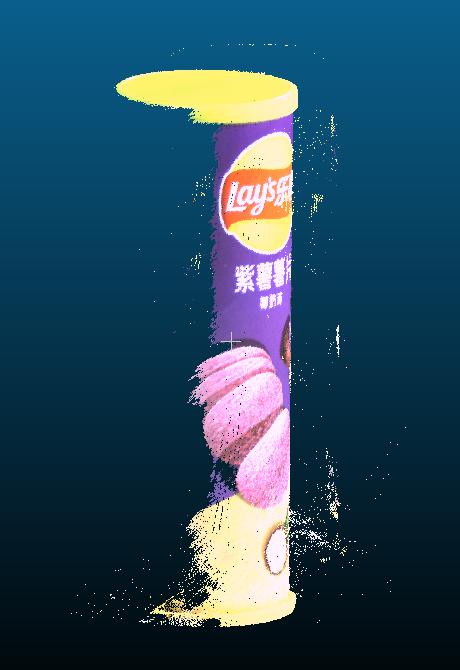} \includegraphics[width=0.45\linewidth]{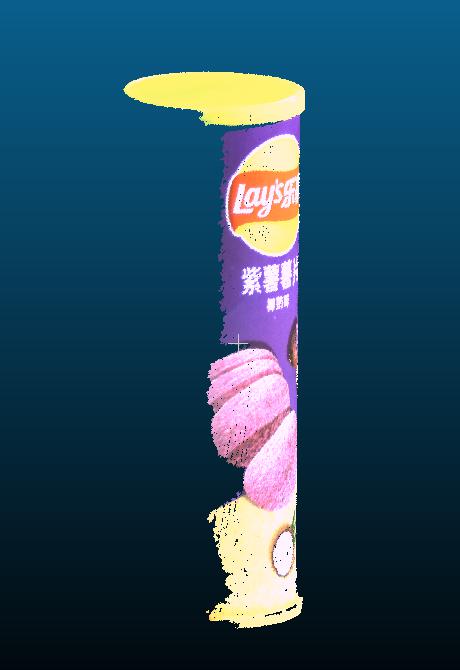}}
    \caption{The left image is the result of segmentation of the original point cloud, the right image shows the filtered point cloud. We can see all noise points have been removed.}
    \label{segment}
\end{figure}
\begin{figure}[htbp]
    \centering
      \subfigure{\includegraphics[width=0.45\linewidth]{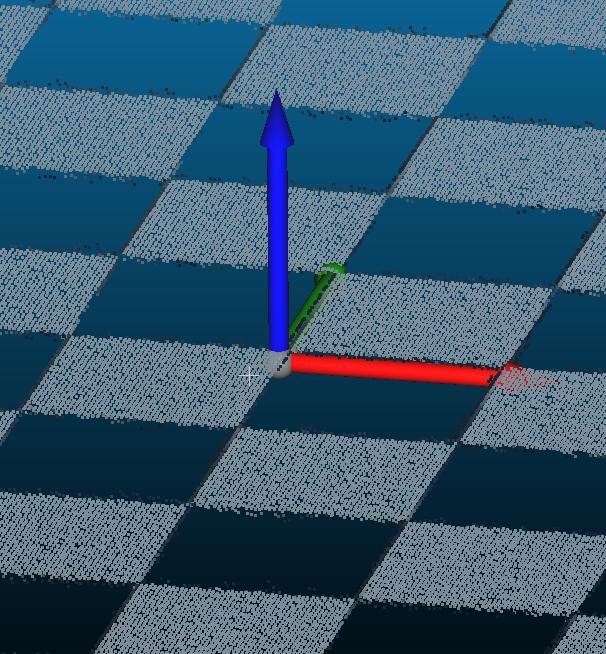} \includegraphics[width=0.45\linewidth]{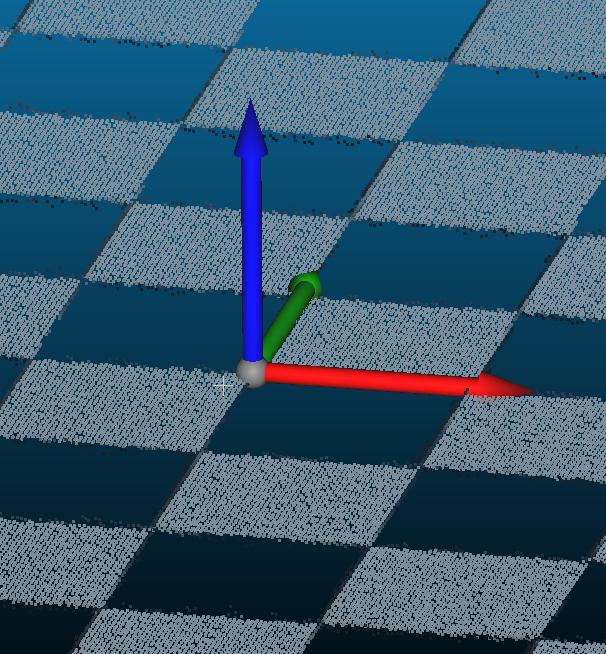}}
    \caption{The left image shows the coordinate origin is below the chessboard plane. The right image shows after scale equalization the coordinate origin is on the chessboard plane.}
    \label{before scale}
\end{figure}
\begin{figure}[htb]
    \centering
    \subfigure{\includegraphics[width=0.45\linewidth]{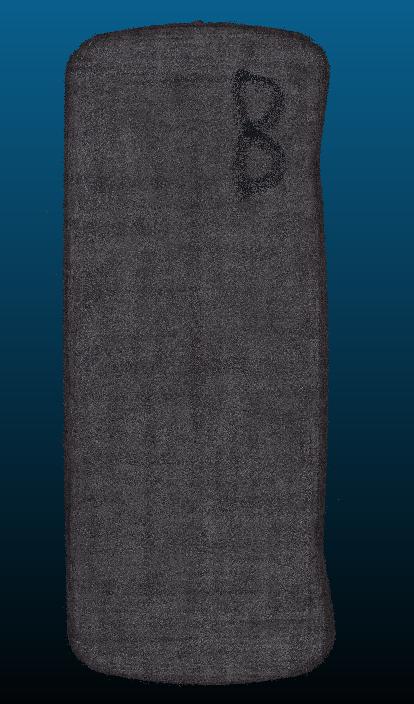} \includegraphics[width=0.45\linewidth]{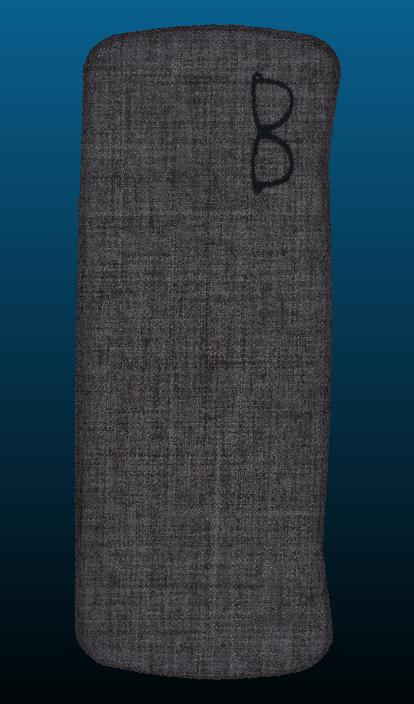}}
    \caption{The left point cloud is generated without scale equalization, the right point cloud is generated with scale equalization, the cloud is much clearer and well-aligned.}
    \label{scale}
\end{figure}
\begin{figure}[ht]
    \centering
    \subfigure{\includegraphics[width=0.45\linewidth]{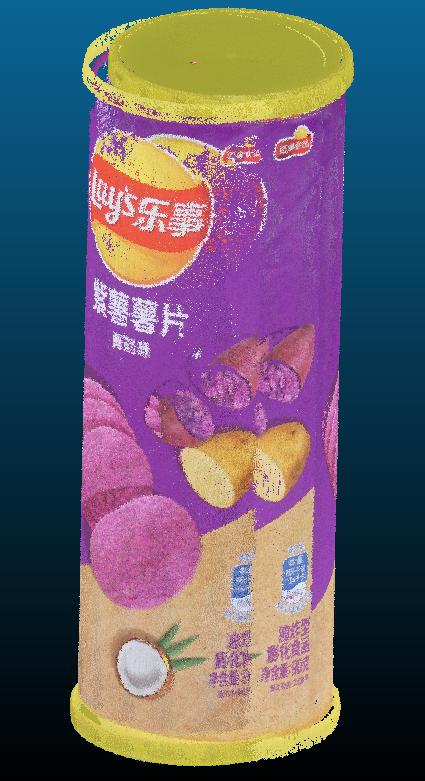} \includegraphics[width=0.45\linewidth]{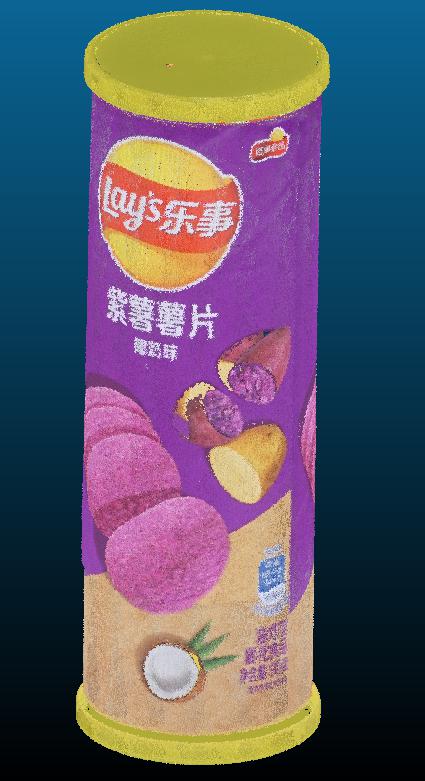}}
    \caption{The left image shows a good initial value for registration. The right image show a good outcome of point cloud registration using Colored-ICP.}
    \label{register}
\end{figure}
As the calibration process is done, it's able to capture the structure of the object. The system runs the following steps:
\begin{itemize}
    \item [1)] Collect data from two cameras in 16 scenarios for chessboard and object separately.
\end{itemize}
\begin{itemize}
    \item [2)] Calculate all poses by using chessboard for each camera position.
\end{itemize}
\begin{itemize}
    \item [3)] Segment raw point cloud by bounding box and remove noise points~\cite{Zhou18}. The details are summarized in Section~\ref{pts_filer}.
\end{itemize}
\begin{itemize}
    \item [4)] Fuse all point clouds to the reference coordinate, and improve the point cloud quality via scale equalization. This part of the details is in Section~\ref{scale_equal}.
\end{itemize}
\begin{itemize}
    \item [5)] Register two point clouds by Colored Point Cloud Registration~\cite{Park17}, one is that the object is put forward, and the other is upside down. This part of the details is in Section~\ref{merge}.
\end{itemize}
\begin{itemize}
    \item [6)] Generation a mesh through Poisson Reconstruction~\cite{Kazhdan06}. 
\end{itemize}
\begin{itemize}
    \item [7)] Dye mesh to improve texture resolution. Section~\ref{impore_texture}.
\end{itemize}

\subsection{Point Cloud Filtering}
\label{pts_filer}
The raw point clouds got from the structured light camera, will get filter by the bounding box which mentioned in Section~\ref{bound} first. This process will get rid of the most of the unusual outliers. In Fig.~\ref{segment}, we visualize qualitative results of this process.

\subsection{Scale Equalization}
\label{scale_equal}

The plane where the chessboard placed is defined as the reference plane, and we move the original point of the coordinate to the center of the chessboard. We compute the transformation $T_{cam\_ref}$ from the camera coordinate to the reference coordinate via solvePnP. We notice the structured light camera has a system error, to be specific, when we transform the chessboard's point cloud to the reference coordinate, it doesn't overlap with the XOY plane but slightly above the XOY plane. The result of the system error is evidenced by Fig.~\ref{scale}. In order to address this issue, we use a scale $\alpha$ to fine tune the point cloud. The depth map got from the structured light camera is represented by $I_{depth}$, the fine tuned depth map is $\Tilde{I}_{depth}=\alpha \cdot I_{depth}$. The scale $\alpha$ is calculated as follow:

\begin{itemize}
    \item [1)] 
    We use all the point on the chessboard plane to estimate the equation of the plane and using RANSAC~\cite{Zhou18} to refine the result. The equation is defined under the reference coordinate system, which is represented as $ax+by+cz+d=0$.
\end{itemize}
\begin{itemize}
    \item [2)] We use the transformation $T_{cam\_ ref}$ to get the coordinates of the camera coordinate's origin in the reference coordinate system $(x',y',z')$.
\end{itemize}
\begin{itemize}
    \item [3)]The distance from the origin to the chessboard plane is represented as $d_0$, and the distance from $(x',y',z')$ to chessboard plane is represented as $d_1$, scale $\alpha=\displaystyle\frac{d_0+d_1}{d_1}$, for each scene $i$ we can get a scale $\alpha_i$. By calculate all 32 scenarios we can get 32 scales. The final scale $\alpha$ is determined by averaging them.
\end{itemize}

For the experiment, we set the scale $\alpha=1.00223$. Fig.~\ref{scale} exhibits the effect of the scale equalization step.
\begin{figure*}[ht]
    \centering
    \subfigure[]{\includegraphics[width=0.195\hsize]{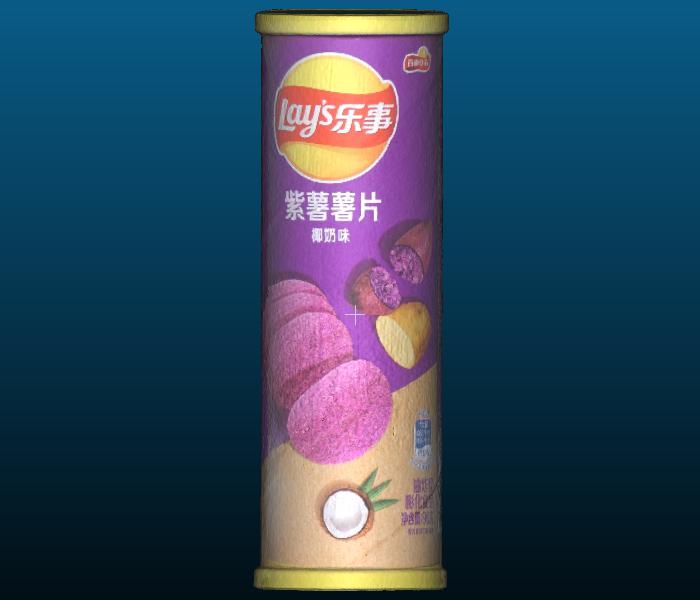}}  
    \subfigure[]{\includegraphics[width=0.195\hsize]{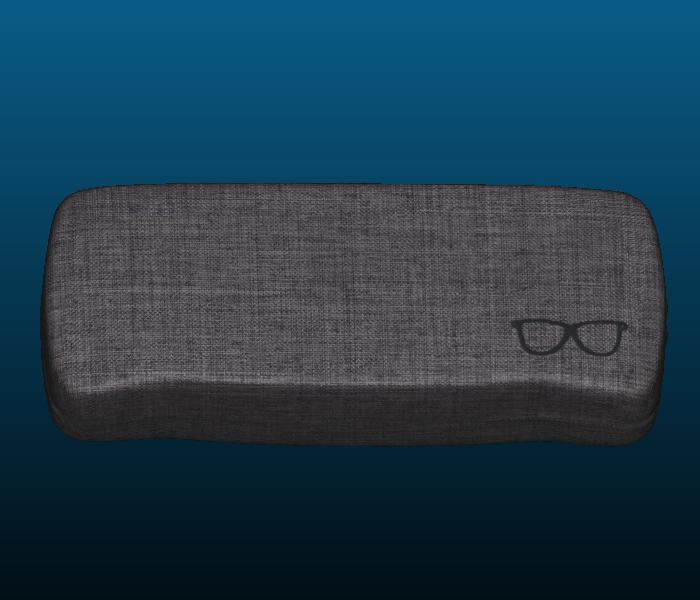}} 
    \subfigure[]{\includegraphics[width=0.195\hsize]{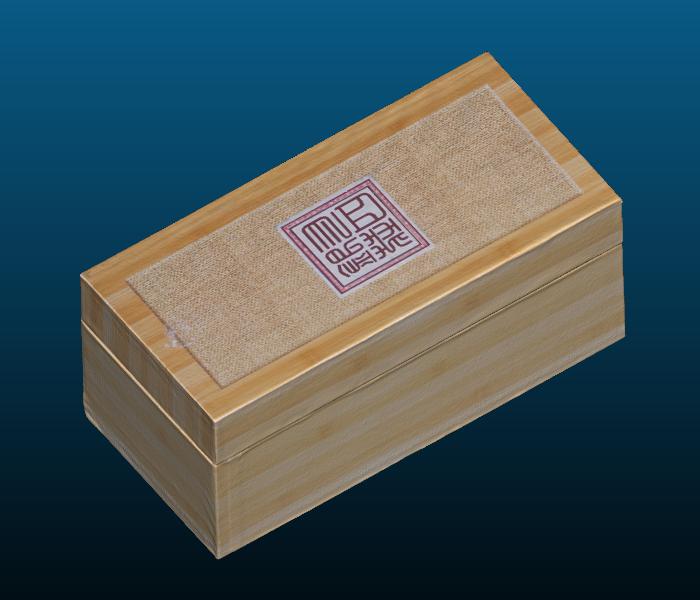}}
    \subfigure[]{\includegraphics[width=0.195\hsize]{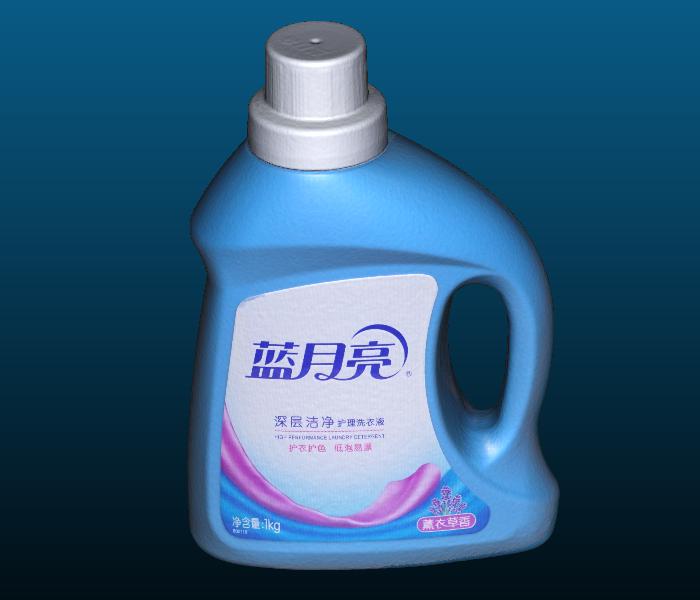}}
    \subfigure[]{\includegraphics[width=0.195\hsize]{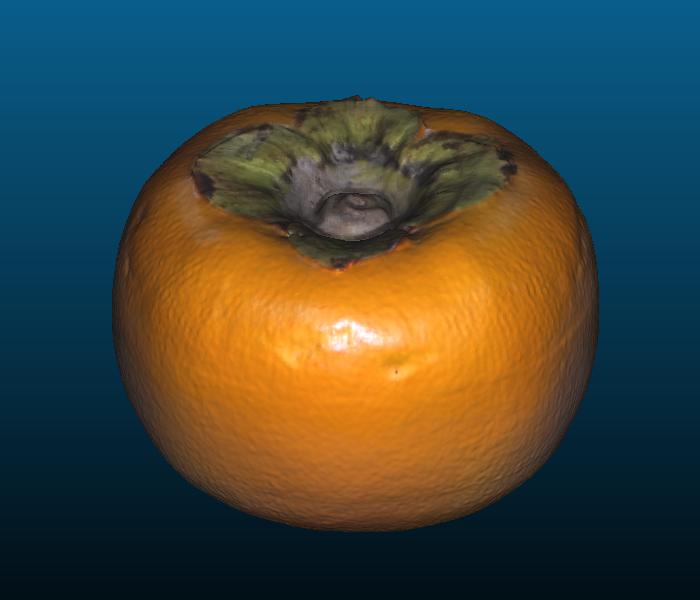}}
    \caption{These are several meshes of our data. Our reconstruction method has good results in shape, color and resolution and the meshes look like real things.}
    \label{mesh}
\end{figure*}
\subsection{Merging Two Point Clouds}
\label{merge}


As we capture the object at two positions, upright and upside down, here we need to concatenate these two point cloud together to get the full structure of the object. Here we choose to use point cloud registration method, such as ICP~\cite{Besl92}, Go-ICP~\cite{Yang15}, Colored-ICP~\cite{Park17} and FGR~\cite{Zhou16}. For many symmetrical objects, it has ambiguity problem when only using the depth information. Letting the texture information involved this issue could be alleviated. In practice, we found the Colored-ICP is the best one to fit this purpose. We know the position of these two point cloud, thus could provide a very good initial value for Colored-ICP. Besides, we only using the middle overlapping part of the object for the registration process. The qualitative result is shown in Fig.~\ref{register}.

\subsection{Improve Texture Quality}
\label{impore_texture}

\begin{figure}[htbp]
    \centering
    \subfigure{\includegraphics[width=0.47\linewidth]{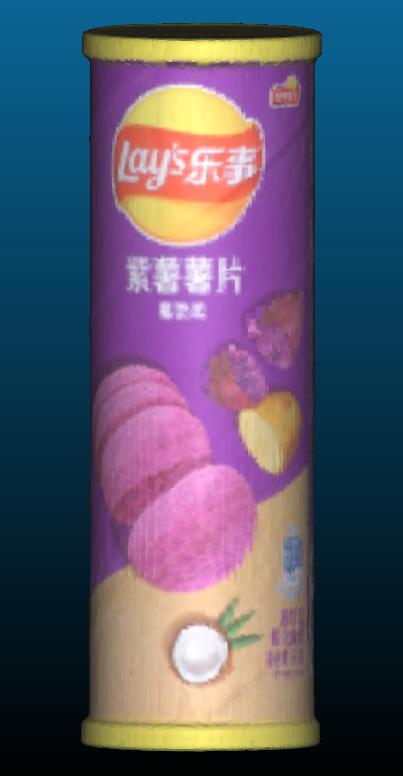} \includegraphics[width=0.47\linewidth]{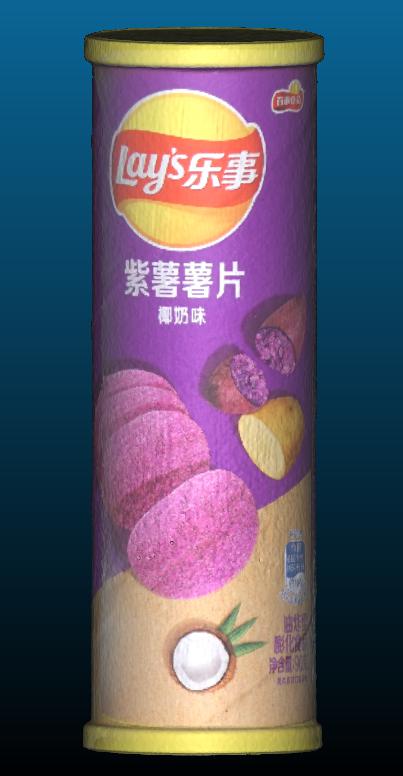}}
    \caption{The left image shows the fuzzy texture of mesh, the right image shows the effect of re dyeing. Through the comparison, we can see that the improvement of the edge is obvious.}
    \label{texture}
\end{figure}

We can get a complete and textured point cloud of the object after the merging process, as shown in the right image in Fig.~\ref{register}. To be notices, the texture details on mesh generated by Poisson reconstruction are blurred, as shown in Fig.~\ref{texture}. In order to improve the quality of the texture, we re-dye the object's color. First, we transform all vertices of the mesh to the camera view. Then using method~\cite{Katz07} to removal the hidden points and keep the visible points. Then we re-project these points to the 2D image plane and use the RGB images which captured by the Sony Camera to get the color at the position of the visible points. This method significantly improves the texture details of object, as shown in Fig.~\ref{texture}.

\subsection{Accuracy}
In order to measure the accuracy of our method, we follow Bigbird's protocol, projecting a representative mesh onto an image from Sony camera(which is only used to get color information). Compared with Bigbird's outcome, we can see that our method could fit the contour of the object better, as shown in Fig.~\ref{projection}. Using our method to generate datasets can have more accurate annotation, which is not trivial in 6D pose estimation.

At the same time, we compare the size of the model with the size of the real object. In order to get accurate measurement result, we choose a metal column. We compare the size of the bounding box of the mesh with its actual physic size. Our reconstruction error is less then 0.2mm, the result is summarized in Table~\ref{tab:01}.
\begin{table}[ht]
\centering
\begin{tabular}{|c|c|c|c|l}
\cline{1-4}
\multirow{2}{*}{direction} & \multicolumn{2}{c|}{length of bounding box(mm)} & \multirow{2}{*}{error(mm)} &  \\ \cline{2-3}
                          & mesh                   & real                   &                            &  \\ \cline{1-4}
x                          & 77.89                  & 77.96                  & 0.07                       &  \\ \cline{1-4}
y                          & 78.09                  & 77.98                  & 0.11                       &  \\ \cline{1-4}
z                          & 85.32                  & 85.48                  & 0.16                       &  \\ \cline{1-4}
\end{tabular}
\caption{Error of reconstruction}
\label{tab:01}
\end{table}

\section{Presentation of Our Data}

\subsection{Mesh data}
At present, we have completed the acquisition of tens of objects(some are shown in Fig.~\ref{mesh}), and will expand the amount of objects to more than 100 in future work.

\subsection{Depth Map and Point Cloud}
While generating mesh, we can generate depth map from each scenarios, as shown in Fig.~\ref{depth}.The highest resolution of those depth cameras is 1.3MP(belongs to Ensenso N35), but its depth map's quality is not the best. Compared with depth maps from several depth sensors, our depth map has higher resolution(6MP) and the edge of object is more sharp. Also we provide high-quality, dense point clouds of objects
(the resolution of point cloud is 0.1 mm), as shown in the right image in Fig.~\ref{register}.

Compared all other sensors' result in Fig.~\ref{depth}, our results is sharp and clean at the edge of the objects. Our data is closer to the data generated by render, the result in Fig.~\ref{depth} confirms our statement, which is reducing the gap between the real data and the synthetic data.
\section{CONCLUSION}

We present a robust, high precision and fast 3D reconstruction method. Using this method, we can provide high resolution meshes, high precision point clouds, high resolution depth maps and high resolution RGB images. In addition, we can generate more accurate ground truth annotation than the existing methods. Our algorithm is fully automatic and efficient, and has the ability to make large datasets. At the same time, the data generated by our algorithm is closer to the data generated by render, which reducing the gap between real data and synthetic data.

In the future, we will use this system to make a large 3D dataset with accurate ground truth to facilitate the 6D pose estimation task.
\clearpage
{\small

\bibliographystyle{unsrt}
}
\end{document}